\documentclass{article}
\usepackage{ijcai19}

\usepackage{times}
\usepackage{soul}
\usepackage{url}
\usepackage[hidelinks]{hyperref}
\usepackage[utf8]{inputenc}
\usepackage[small]{caption}
\usepackage{graphicx}
\usepackage{amsmath}
\usepackage{booktabs}
\usepackage{float} 
\usepackage{enumitem}
\usepackage[misc]{ifsym}
\usepackage{dsfont}
\usepackage{amsfonts}
\usepackage{mathtools}
\newcommand{\norm}[1]{\left\| #1 \right\|} 
\usepackage{algorithm}
\usepackage[noend]{algpseudocode}
\algdef{SE}[DOWHILE]{Do}{doWhile}{\algorithmicdo}[1]{\algorithmicwhile\ #1}
\usepackage{subcaption}
\usepackage{multirow}

\usepackage[utf8]{inputenc}
\usepackage{pgfplots}
\pgfplotsset{compat=1.13}
\newlength\figureheight
\newlength\figurewidth
\title{Matrix Completion in the Unit Hypercube via Structured Matrix Factorization}

\author{
Emanuele Bugliarello$^1$\footnote{Majority of work done while at Technicolor AI Lab.}\and
Swayambhoo Jain$^2$\And
Vineeth Rakesh$^2$\\
\affiliations
$^1$Tokyo Institute of Technology\\
$^2$Technicolor AI Lab\\
\emails
emanuele.bugliarello@nlp.c.titech.ac.jp,
\{Swayambhoo.Jain, Vineeth.Mohan\}@technicolor.com
}

\begin{document}

\maketitle

\begin{abstract}

Several complex tasks that arise in organizations can be simplified by mapping them into a matrix completion problem. 
In this paper, we address a key challenge faced by our company: predicting the efficiency of artists in rendering visual effects (VFX) in film shots. 
We tackle this challenge by using a two-fold approach: first, we transform this task into a constrained matrix completion problem with \textit{entries bounded in the unit interval $[0,1]$}; second, we propose two novel matrix factorization models that leverage our knowledge of the VFX environment. 
Our first approach, \textit{expertise matrix factorization} (EMF), is an interpretable method that structures the latent factors as weighted user-item interplay. 
The second one, \textit{survival matrix factorization} (SMF), is instead a probabilistic model for the underlying process defining employees’ efficiencies.
We show the effectiveness of our proposed models by extensive numerical tests on our VFX dataset and two additional datasets with values that are also bounded in the $[0,1]$ interval.
\end{abstract}

\section{Introduction} \label{sec:intro}

\begin{figure}
	\includegraphics[width=0.45\textwidth, trim={2.5cm 3.2cm 2.6cm 3.2cm}, clip]{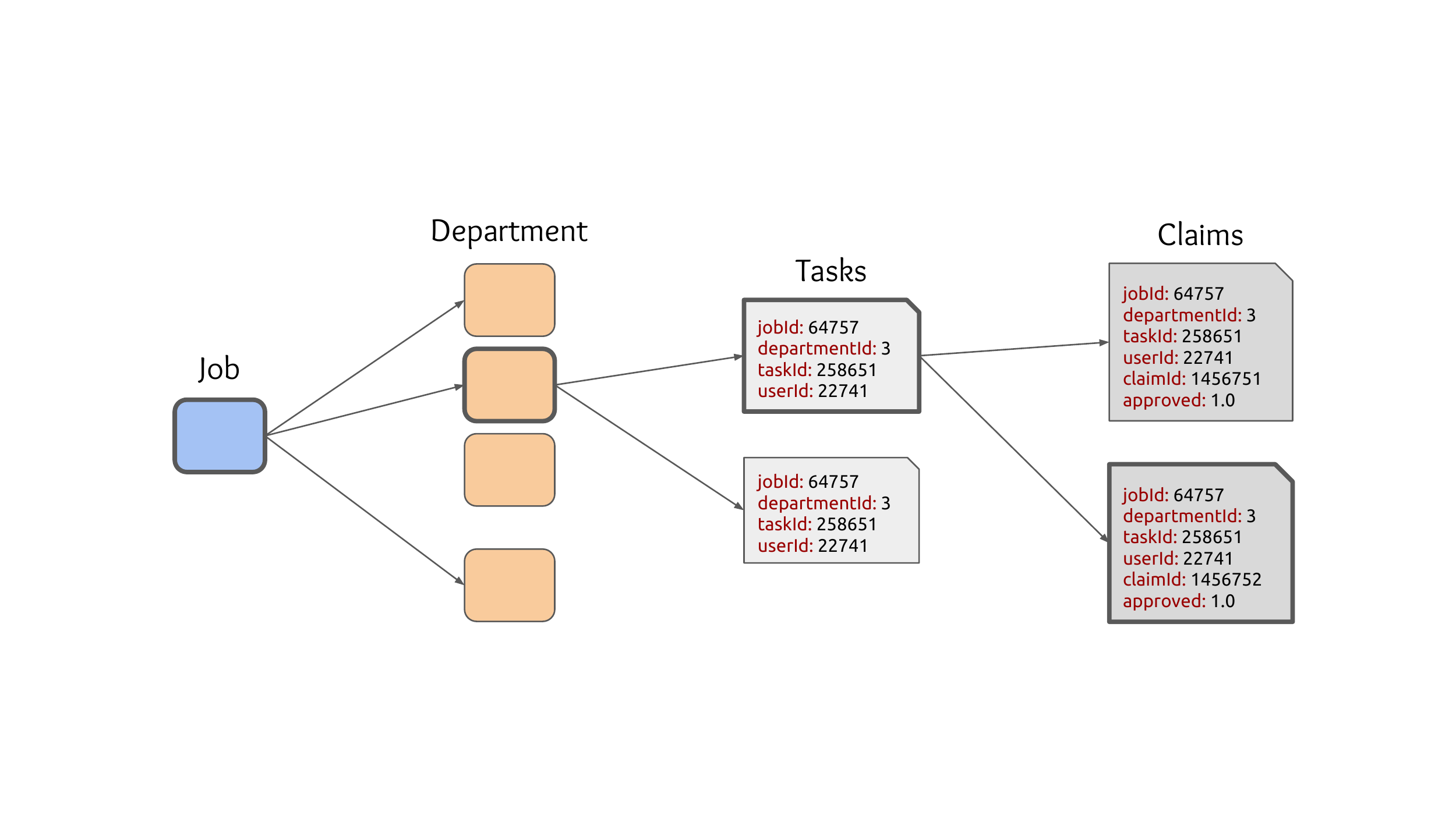}
	\caption{Hierarchical task allocation.}
	\label{fig:organization}
\end{figure}

A variety of complex applications that emerge in several organizations can be translated to matrix completion problems. 
In this study, we address a key challenge faced by our company: predicting the efficiency of artists in rendering visual effects (VFX) in film shots. 
Effectively solving this problem is of crucial importance as project managers often rely on competency metrics, such as employee’s efficiency, to best control and adjust the available assets.
Figure~\ref{fig:organization} illustrates the hierarchical relationship between artists (i.e., employees) and job assignments in a VFX production framework.
Here, a \textit{job} for the organization just consists of a series of contributions from different \textit{departments} and possibly involves numerous \textit{employees}. 
Each department is led by a \textit{manager} who divides the work in her department into multiple \textit{tasks}, and each task is assigned to a single employee. 
Employees work on these tasks and submit partial results called \textit{claims}. 
After assessing the quality of a claim, the manager decides whether to approve it or not. 
Hence, our goal is to exploit the latent attributes of each employee that explain their efficiency across various tasks to ensure products can be delivered on time.

The key contributions of our work are as follows: (i) transforming industrial setups into a sparse real-valued matrix with entries lying in the $[0,1]$ interval (e.g., efficiencies of employees in departments), and (ii) proposing two novel structured matrix factorization (MF) models to effectively solve the resulting matrix completion problem. While our methods are inspired by the hierarchical task allocation shown in Figure~\ref{fig:organization}, their formulation is very general and can easily be applied to a variety of scenarios. In this paper, in particular, we consider two additional applications: (a) recommendation of apps to the users of over-the-top (OTT) streaming devices, and (b) recommendation in online advertisement placement to maximize click-through rate (CTR). The details of these applications will be explained in later sections.

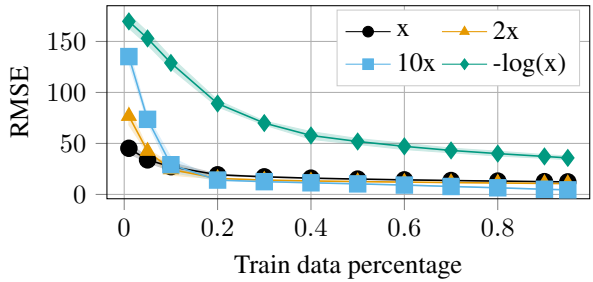
\begin{figure}
	\setlength\figureheight{4.15cm}
	\setlength\figurewidth{8cm}
\begin{tikzpicture}

\definecolor{color0}{rgb}{0.9,0.6,0}
\definecolor{color1}{rgb}{0.35,0.7,0.9}
\definecolor{color2}{rgb}{0,0.6,0.5}

\begin{axis}[
height=\figureheight,
legend cell align={left},
legend entries={{x},{2x},{10x},{-log(x)}},
legend style={draw=white!80.0!black},
legend columns=2,
tick align=outside,
tick pos=left,
width=\figurewidth,
x grid style={white!69.01960784313725!black},
xlabel={Train data percentage},
xmajorgrids,
xmin=-0.037, xmax=0.997,
y grid style={white!69.01960784313725!black},
ylabel={RMSE},
ymajorgrids,
ymin=-4.24850735028519, ymax=185.041345670371
]
\addlegendimage{mark=*, black}
\addlegendimage{mark=triangle*, color0}
\addlegendimage{mark=square*, color1}
\addlegendimage{mark=diamond*, color2}
\path [draw=black, fill=black, opacity=0.2] (axis cs:0.01,50.4019746545173)
--(axis cs:0.01,40.0775054502982)
--(axis cs:0.05,31.1934313517793)
--(axis cs:0.1,23.1728124496647)
--(axis cs:0.2,18.5510831192238)
--(axis cs:0.3,16.9071190997121)
--(axis cs:0.4,15.7042406522212)
--(axis cs:0.5,14.8822723678844)
--(axis cs:0.6,14.1978033653298)
--(axis cs:0.7,13.5426360910488)
--(axis cs:0.8,12.9036857315682)
--(axis cs:0.9,12.4316168445065)
--(axis cs:0.95,12.1954881835645)
--(axis cs:0.95,12.4167242831556)
--(axis cs:0.95,12.4167242831556)
--(axis cs:0.9,12.6435552675366)
--(axis cs:0.8,13.2391245527068)
--(axis cs:0.7,13.7057841655996)
--(axis cs:0.6,14.3676603462391)
--(axis cs:0.5,15.2040507000044)
--(axis cs:0.4,16.1561651387489)
--(axis cs:0.3,17.5846673795405)
--(axis cs:0.2,19.9526577320709)
--(axis cs:0.1,31.2470192295076)
--(axis cs:0.05,36.5159540493489)
--(axis cs:0.01,50.4019746545173)
--cycle;

\path [draw=color0, fill=color0, opacity=0.2] (axis cs:0.01,85.4549349374903)
--(axis cs:0.01,68.1015391003121)
--(axis cs:0.05,40.3251444168501)
--(axis cs:0.1,18.8190730322379)
--(axis cs:0.2,15.1502541838529)
--(axis cs:0.3,13.7286828630328)
--(axis cs:0.4,12.9729368251608)
--(axis cs:0.5,12.5214716240709)
--(axis cs:0.6,12.0377402545232)
--(axis cs:0.7,11.531186504546)
--(axis cs:0.8,11.099536871827)
--(axis cs:0.9,10.6956356877312)
--(axis cs:0.95,10.517731279611)
--(axis cs:0.95,10.6491166345309)
--(axis cs:0.95,10.6491166345309)
--(axis cs:0.9,10.840193471182)
--(axis cs:0.8,11.318679967695)
--(axis cs:0.7,11.6951130418537)
--(axis cs:0.6,12.1122878319241)
--(axis cs:0.5,12.6843434513657)
--(axis cs:0.4,13.3993285836927)
--(axis cs:0.3,14.3640316886779)
--(axis cs:0.2,16.200356942968)
--(axis cs:0.1,30.0192600639584)
--(axis cs:0.05,44.5880062418685)
--(axis cs:0.01,85.4549349374903)
--cycle;

\path [draw=color1, fill=color1, opacity=0.2] (axis cs:0.01,141.927640175029)
--(axis cs:0.01,128.37852961241)
--(axis cs:0.05,70.440380964053)
--(axis cs:0.1,21.7040447551706)
--(axis cs:0.2,13.6162420055216)
--(axis cs:0.3,12.2986419273347)
--(axis cs:0.4,11.0744457691773)
--(axis cs:0.5,10.2674575216515)
--(axis cs:0.6,9.00912611642283)
--(axis cs:0.7,7.51586793922855)
--(axis cs:0.8,6.40107865345505)
--(axis cs:0.9,4.86890413713299)
--(axis cs:0.95,4.35557687792645)
--(axis cs:0.95,4.57591962557599)
--(axis cs:0.95,4.57591962557599)
--(axis cs:0.9,5.25820063127197)
--(axis cs:0.8,6.58460150734896)
--(axis cs:0.7,8.08227531631248)
--(axis cs:0.6,9.3395225884757)
--(axis cs:0.5,10.4058688676292)
--(axis cs:0.4,11.730734117698)
--(axis cs:0.3,12.7398577176746)
--(axis cs:0.2,14.5325977763351)
--(axis cs:0.1,36.5355718898606)
--(axis cs:0.05,76.8034652362497)
--(axis cs:0.01,141.927640175029)
--cycle;

\path [draw=color2, fill=color2, opacity=0.2] (axis cs:0.01,176.437261442159)
--(axis cs:0.01,162.935480403261)
--(axis cs:0.05,146.537346901007)
--(axis cs:0.1,123.133762611724)
--(axis cs:0.2,85.7882641392169)
--(axis cs:0.3,68.02348854791)
--(axis cs:0.4,54.0289679313136)
--(axis cs:0.5,49.0257518482452)
--(axis cs:0.6,44.0255469617617)
--(axis cs:0.7,40.8652107579906)
--(axis cs:0.8,37.6602555569895)
--(axis cs:0.9,35.4159779331552)
--(axis cs:0.95,34.0052365735933)
--(axis cs:0.95,37.5976871003819)
--(axis cs:0.95,37.5976871003819)
--(axis cs:0.9,39.0881968952453)
--(axis cs:0.8,42.3775120652047)
--(axis cs:0.7,45.3887605330126)
--(axis cs:0.6,50.2434333183992)
--(axis cs:0.5,54.7745786382272)
--(axis cs:0.4,61.6148317432723)
--(axis cs:0.3,71.9017102649099)
--(axis cs:0.2,92.3953602671135)
--(axis cs:0.1,134.750922285959)
--(axis cs:0.05,159.370317626793)
--(axis cs:0.01,176.437261442159)
--cycle;

\addplot [semithick, black, mark=*, mark size=3, mark options={solid}]
table [row sep=\\]{%
0.01	45.2397400524078 \\
0.05	33.8546927005641 \\
0.1	27.2099158395862 \\
0.2	19.2518704256474 \\
0.3	17.2458932396263 \\
0.4	15.930202895485 \\
0.5	15.0431615339444 \\
0.6	14.2827318557845 \\
0.7	13.6242101283242 \\
0.8	13.0714051421375 \\
0.9	12.5375860560216 \\
0.95	12.30610623336 \\
};
\addplot [semithick, color0, mark=triangle*, mark size=3, mark options={solid}]
table [row sep=\\]{%
0.01	76.7782370189012 \\
0.05	42.4565753293593 \\
0.1	24.4191665480981 \\
0.2	15.6753055634104 \\
0.3	14.0463572758554 \\
0.4	13.1861327044268 \\
0.5	12.6029075377183 \\
0.6	12.0750140432236 \\
0.7	11.6131497731999 \\
0.8	11.209108419761 \\
0.9	10.7679145794566 \\
0.95	10.5834239570709 \\
};
\addplot [semithick, color1, mark=square*, mark size=3, mark options={solid}]
table [row sep=\\]{%
0.01	135.15308489372 \\
0.05	73.6219231001513 \\
0.1	29.1198083225156 \\
0.2	14.0744198909284 \\
0.3	12.5192498225046 \\
0.4	11.4025899434377 \\
0.5	10.3366631946404 \\
0.6	9.17432435244927 \\
0.7	7.79907162777052 \\
0.8	6.49284008040201 \\
0.9	5.06355238420248 \\
0.95	4.46574825175122 \\
};
\addplot [semithick, color2, mark=diamond*, mark size=3, mark options={solid}]
table [row sep=\\]{%
0.01	169.68637092271 \\
0.05	152.9538322639 \\
0.1	128.942342448841 \\
0.2	89.0918122031652 \\
0.3	69.9625994064099 \\
0.4	57.821899837293 \\
0.5	51.9001652432362 \\
0.6	47.1344901400805 \\
0.7	43.1269856455016 \\
0.8	40.0188838110971 \\
0.9	37.2520874142003 \\
0.95	35.8014618369876 \\
};
\end{axis}

\end{tikzpicture}
	\vspace{-0.3cm}
	\caption{Prediction errors (mean $\pm$ standard error) on five $40\times30$ random matrices with entries in $[0, 1]$ when applying monotonic mappings at various training data percentages. These transformations do not ease the matrix factorization problem.}
	\label{fig:monotonic}
\end{figure}

Although conventional matrix factorization algorithms have been proved effective to solve the matrix completion problem, recent studies show that constrained MF techniques outperform them on a variety of datasets, including MovieLens, Jester and BookCrossing~\cite{Fang:2017:IBM:3172077.3172117,jawanpuria2018unified}. Moreover, MF approaches tend to produce unstable and out-of-range predictions as the sparsity of the matrix increases~\cite{jiang2018magnitude}.
While most recommender systems have entries bounded within a range of possible rating values (e.g., one to five in a five-star rating system), our data lies on the unit interval $[0, 1]$.
While we could convert our data to an ordinal space, such as ratings between one and five, as shown in Figure~\ref{fig:monotonic}, any monotonic transformation mapping the unit interval to a larger one does not simplify the prediction problem.
Therefore, we propose two novel approaches for this scenario: \textit{expertise matrix factorization} (EMF), in which entries are approximated by a weighted correlation between user and item latent factors (i.e., the low-rank features), and \textit{survival matrix factorization} (SMF), which uses a probabilistic model to capture the latent factors by modeling the procedure resulting in the average efficiency of an employee in a department. Both our approaches are structured matrix factorization methods with constrains derived by modeling the presented setup. 

We validate the effectiveness of the proposed models using three datasets: (a) a proprietary dataset of VFX efficiencies, (b) a proprietary dataset of app usage in OTT device, and (c) a public dataset of CTR released by Outbrain. Using a rigorous series of numerical experiments, we show that the proposed models outperform popular MF techniques on scores such as RMSE, MAE, Precision@N and Recall@N.

 \section{Related Work}\label{sec:related}
The allocation of resources is a relevant problem in Business Process Management to improve the performance of an organization~\cite{dumas2013fundamentals}. Recently, different approaches in Process Mining~\cite{van2011process} have been proposed to extract useful knowledge from historical data~\cite{arias2018human}. However, previous studies focused on particular process cases~\cite{huang2012task,conforti2015recommendation}. In response, we investigate a generic framework that can be mapped to a variety of organizations.
Our proposed approaches are inspired by a series of work on low-rank matrix completion, and their applications to recommendation systems~\cite{hu2008collaborative,candes2009exact,candes2010matrix,koren2009matrix,funk2011netflix,jain2013low}.
A class of approaches suitable for our task consists of bounded matrix factorization methods. Non-negative matrix factorization (NMF) is the most popular method in this class, and it only provides a lower bound of $0$ for the predicted values. A more general approach is given by bounded matrix factorization (BMF)~\cite{kannan2014bounded}: a low-rank approximation that leverages the fact that all ratings in a recommender system are bounded within a range $[x_{min}, x_{max}]$
Our first approach, expertise matrix factorization, is closely related to these approaches and to matrix completion models with additional structure on the latent factors~\cite{soni2016noisy,hoyer2004non,aharon2006k,kannan2014bounded}. However, our approach departs from them in its unique structure given by the unit interval constraint in our setup. To the best of our knowledge, this structure has not been specifically tackled in the existing literature of MF with explicit data. MF methods for implicit feedback data~\cite{hu2008collaborative,johnson2014logistic} work on the same interval, but with the major difference of relying on just matrix entries that are binary values.
Our second approach, survival matrix factorization, is instead related to probabilistic matrix factorization methods~\cite{mnih2008probabilistic,Salakhutdinov:2008:BPM:1390156.1390267}. While they also interpret matrix entries as probabilities, our method specifically models the end-to-end approval process in its formulation.\\
 
\noindent
\textbf{Notations}. 
In this paper, all vectors are represented by bold lowercase letters and are column vectors (e.g., \textbf{p}, \textbf{q}). All matrices are represented by bold uppercase letters (e.g., \textbf{P}, \textbf{Q}). For a given matrix \textbf{A}, $\textbf{a}_i$ denotes its $i$-th row as a column vector and $a_{ij}$ denotes its entry in row $i$ and column $j$.

\section{Problem Formulation}\label{sec:problem}

As described in Section~\ref{sec:intro}, our proposed models are inspired by the framework depicted in Figure~\ref{fig:organization}. This pipeline is motivated by the concrete application scenario of work allocation in movie production at Technicolor, where artists (employees) render visual effects in film shots across various disciplines (departments).
Nonetheless, this framework is very general and other production pipelines can easily be mapped into this setting.
For instance, if an organization logs employees' performance on pieces of work in a $100$-point scale, then each score can be mapped to an approved or rejected claim by defining a threshold value of work quality (e.g., $90/100$).

Considering the work allocation scenario, a natural competency metric for employees' efficiencies is given by the ratio of number of claims accepted to the total number of claims submitted by the employee in the department. 
An accepted claim counts towards the employee's overall performance. In case of rejection, the employee submits a new claim when they believe it will meet the quality required. The latter scenario clearly results in a loss in performance for the entire organization as the employee might need to start over or the manager may decide to designate another employee.
Let $N_{dn}$ be the total number of claims by employee $d$ in department $n$, and $a_{dn}^{(i)} $ denote whether the $i$-th claim by employee $d$ in department $n$ was accepted ($1$) or rejected ($0$).
We then define the efficiency of employee $d$ in department $n$, $x_{dn}$, as follows:
\begin{equation} \label{eqn:tch_efficiency}
\displaystyle x_{dn} = \frac{\sum_{i=1}^{N_{dn}}a_{dn}^{(i)}}{N_{dn}}.
\end{equation}
It is straightforward to see that $x_{dn} \in [0,1]$.

The goal of our study is to predict the efficiencies of employees in each department. 
We address this challenge by building an efficiency matrix $\mathbf{X} \in \mathbb{R}^{D \times N}$ whose entry $x_{dn} \in [0,1]$ denotes the efficiency of employee $d$ in department $n$. Only a few entries of this matrix, corresponding to the observation index set $\Omega \subset [D] \times [N]$, are known since most employees only work in a few departments throughout their careers.
Hence, our goal of predicting employees' efficiencies now reduces to predicting the missing entries of $\mathbf{X}$.

\newpage

\section{Proposed Models}\label{sec:models}

Our first proposed model, \textit{expertise matrix factorization}, builds on low-rank models and introduces additional structure on the factors $\mathbf{W}$ and $\mathbf{Z}$ so that the entries of the resulting matrix $\mathbf{W} \cdot \mathbf{Z}^T$ lie in unit interval. 
Our second model, \textit{survival matrix factorization}, instead follows a probabilistic modeling of the claim approval process defining the efficiency matrix.

\subsection{Expertise Matrix Factorization}
In this first approach, we model the efficiency matrix as a low-rank matrix $\textbf{X} \approx \textbf{W}\cdot\textbf{Z}^T$. 
In our setup, we can think of the latent factors as the set of \textit{skills} or \textit{expertise} required to work in a given department. 
In particular, we assume each employee's latent factors (skills) to range from $0$ to $1$, while each department's latent factors to be non-negative and to sum to $1$.
On the one hand, employees have a certain level in each skill, where $0$ indicates no ability and $1$ proficiency. On the other hand, each department is assumed to have a given proportion of skills required to complete tasks in it.

We assume a low-rank model based on the intuitive reasoning that a small number of skills may be required to complete tasks across different departments. 
With this model, the efficiency of employee $d$ in department $n$ is approximated by a weighted sum of the employee's skills, where the weights are given by the importance of each skill in the department. The structure in the latent vectors ensures that efficiency values under this model lie in $[0,1]$. 
We further extend this model to accommodate user biases, representing the minimum proficiency of each employee in every skill. Note that it would be inconsistent to add a bias term for departments as not every skill is supposed to be useful in each discipline.

The resulting optimization problem is as follows:
\begin{equation} \label{eqn:loss_constrained_bias}
\begin{aligned}
\underset{\textbf{W} \ge 0,\textbf{Z} \ge 0, \boldsymbol{\beta} \ge 0}{\text{min}} & \hspace{1em} \frac{1}{2 |\Omega|}  \sum_{(d,n)\in\Omega}  \left[x_{dn} - \left(\beta_d + \textbf{w}^T_d\cdot\textbf{z}_n\right)\right]^2 \\
& ~~~~~~ + \frac{\lambda_u}{2}\left( \norm{\textbf{W}}_F^2 + \norm{\boldsymbol{\beta}}_2^2 \right) + \frac{\lambda_i}{2}\norm{\textbf{Z}}_F^2\\
\text{subject to} \hspace{0.8em} & ~~~~ \beta_d + w_{dk} \leq 1,~~ \text{for } (d,k)\in [D] \times [K]\\
& ~~~ \sum_{k=1}^Kz_{nk} = 1, ~~ \text{for } n \in [N]
\end{aligned}
\end{equation}
where $\boldsymbol{\beta}\in\mathds{R}^D$ is the vector of employees' biases, $\Omega$ is the set of observed entries and $|\Omega|$ is its size, $\lambda_u$ and $\lambda_i$ control the extent of regularization in the users and items parameters, and $||\cdot||_F$ denotes the Frobenius norm. 
The first constraint $\beta_d + w_{dk} \leq 1$ enforces predictions in $[0,1]$. In fact, $\hat{x}_{dn} = \beta_d + \mathbf{w}_d^T\cdot \mathbf{z}_n = \left(\beta_d \mathbf{1} + \mathbf{w}_d\right)^T\cdot \mathbf{z}_n \in [0,1]$ if $\beta_d + w_{dk} \leq 1$. 

The structure on the factor $\mathbf{Z}$ has also been considered in blind de-mixing applications~\cite{lin2015identifiability,fu2018identifiability}, which, however, do not deal with missing entries and do not impose any structure on the factor $\mathbf{W}$. 
The optimization problem in Equation \eqref{eqn:loss_constrained_bias} is jointly non-convex in $\mathbf{W}, \boldsymbol{\beta}$ and $\mathbf{Z}$. However, for fixed values of $\mathbf{W}$ and $\boldsymbol{\beta}$, the problem is a convex quadratic program in $\mathbf{Z}$, and for fixed value $\mathbf{Z}$ the problem is a convex quadratic program in $\mathbf{W}$ and $\boldsymbol{\beta}$ \cite{boyd2004convex}. 
Hence, we rely on an algorithm based on alternating minimization to solve this problem, first solving for \textbf{W} and $\boldsymbol{\beta}$, and then for \textbf{Z}.
In particular, the former sub-problem can be solved by any quadratic program solver, while the latter by a projected gradient algorithm that maps each row of $\mathbf{Z}$ onto the probability simplex.

\subsection{Survival Matrix Factorization}\label{smf_algo}
\begin{figure}
	\includegraphics[width=0.5\textwidth, trim={1cm 0cm 0cm 0cm}, clip]{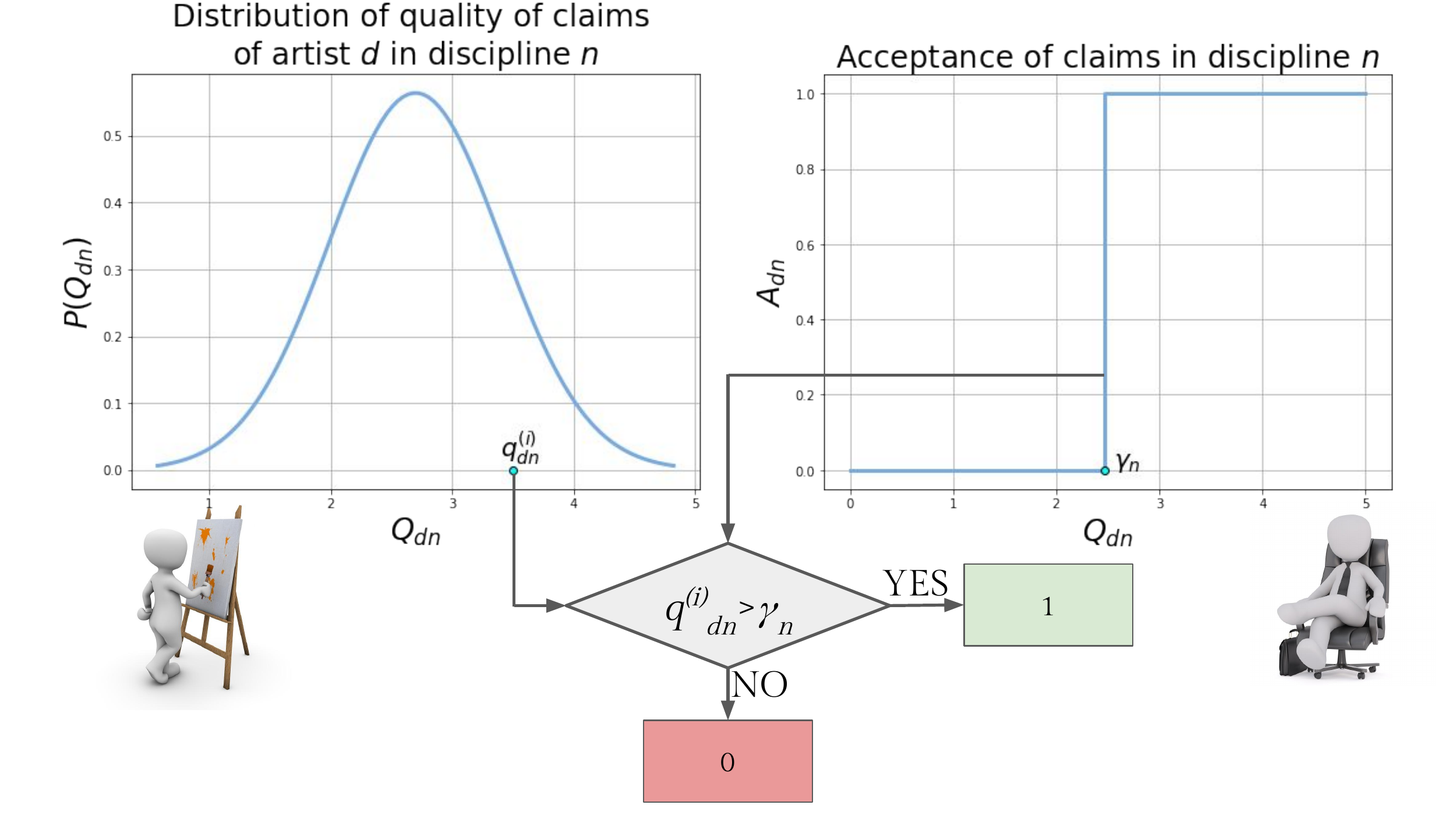}
	\vspace{-0.6cm}
	\caption{Claims acceptance survival model.}
	\label{fig:survival_model}
\end{figure}
In this approach, we aim to derive a probabilistic model for the underlying process resulting in efficiency values.
Recall from Equation~\eqref{eqn:tch_efficiency} that each entry $x_{dn}$ in the efficiency matrix \textbf{X} is defined as the empirical average efficiency of employee $d$ in department $n$. Each claim submitted by an employee can be viewed as an i.i.d. sample from a Bernoulli random variable $A_{dn}$, and, by the Weak Law of Large Numbers, the matrix entry $x_{dn}$ tends almost surely to the mean value of $A_{dn}$, which is equal to $\mathds{P}(A_{dn} = 1)$. 

Having introduced this probability framework, in order to define $\mathds{P}\left( A_{dn} = 1 \right)$, we model the process of claims acceptance. Claims represent pieces of work that employees deliver to managers, who assess the quality of each claim and accept them only if they are good enough.
First, we assume employee $d$ to submit claims to department $n$ with a certain quality distribution. That is, each claim $i$ has quality $Q_{dn}$, a random variable sampled from an underlying probability distribution of quality of work produced by employee $d$ in department $n$. 
Second, we model the manager of department $n$ by a single parameter, $\gamma_{n}$, representing their quality threshold: a claim in department $n$ is only accepted if its quality is higher than the manager's threshold (as depicted in Figure \ref{fig:survival_model}).

Hence, each entry $x_{dn}$ is approximated by:
\begin{equation} \label{eq:approx_smf}
\begin{aligned}
x_{dn} &\approx \mathds{P}\left( A_{dn} = 1 \right) = \mathds{P}\left( Q_{dn} > \gamma_{n} \right) \\&= S_{Q_{dn}}\left( \gamma_{n} \right) = \int_{\gamma_{n}}^{+\infty}f_{Q_{dn}}(u)~du,
\end{aligned}
\end{equation}
where $S_{Q_{dn}}\left(\gamma_{n} \right)$ is the survival function, also known as complementary cumulative distribution function, of $Q_{dn}$ at $\gamma_{n}$, and $f_{Q_{dn}}$ is its probability distribution. 
Note also that each manager is modeled by a single variable and they are then assumed to adhere to equal opportunity laws, accepting claims with no discrimination of race, color, sex, sexual orientation or any other status of the artists.

In this paper, we assume the employees' quality probability distributions to be normal and that the variance of the quality of the claims is the same over all employees and departments:
\begin{equation}
f_{Q_{dn}}\left(x\middle\vert \mu_{dn}, \sigma^2\right) = \frac{1}{\sqrt{2\pi\sigma^2}}e^{-\frac{\left(x - \mu_{dn}\right)^2}{2\sigma^2}},
\end{equation}
where $\mu_{dn}$ is the mean of the distribution of quality of claims of employee $d$ in department $n$ and $\sigma^2$ is its variance.

Furthermore, we try to explain the means $\mu_{dn}$ by characterizing both artists and disciplines with vectors of factors inferred from patterns in the data such that employee-department interactions are modeled as inner products in that space; i.e., $\mu_{dn}\approx\textbf{w}_d^T\cdot\textbf{z}_n$. That is:

\begin{equation}
\hat{x}_{dn} = \displaystyle\int_{\gamma_{n}}^{+\infty} \frac{1}{\sqrt{2\pi\sigma^2}}e^{-\frac{\left(u - \textbf{w}_d^T\cdot\textbf{z}_n \right)^2}{2\sigma^2}} du.
\end{equation}

\noindent
The optimization problem we aim to solve is then given by:
\begin{equation} \label{eqn:smf_loss}
\begin{aligned}
\underset{\textbf{W},\textbf{Z}, \boldsymbol{\gamma}, \sigma}{\text{min}}~ \frac{1}{2|\Omega|}\sum_{(d,n)\in\Omega}\left[x_{dn} - \hat{x}_{dn}  \right]^2 + \frac{\lambda_u}{2}\norm{\textbf{W}}_F^2 + \frac{\lambda_i}{2}\norm{\textbf{Z}}_F^2.
\end{aligned}
\end{equation}

\noindent
The objective function in the above problem is smooth in $\textbf{W},\textbf{Z}, \boldsymbol{\gamma}$ and $\sigma$, and stochastic gradient descent can be used to solve it. We use Leibniz's rule for differentiation under the integral sign to obtain the closed form expressions of the gradients with respect to $\textbf{W},\textbf{Z} \text{ and } \sigma$, and the fundamental theorem of calculus for the gradient with respect to $\boldsymbol{\gamma}$.

\section{Experimental Evaluations}\label{sec:results}
In this section, we evaluate the empirical performance of our models on (a) a dataset of VFX rendering, (b) a dataset of OTT streams from Technicolor, and (c) a public dataset of click-through rates (CTR) in online advertising by Outbrain.
Experimental results on public data are available on GitHub~\footnote{URL: \url{https://github.com/e-bug/unit-mf}.}.

\subsection{Methodology}
\subsubsection{Evaluation Metrics}
The quality of a recommendation algorithm can be evaluated using different types of metrics. We use RMSE and MAE as statistical accuracy metrics, while Precision@N and Recall@N as decision support metrics, for $N\in\{2,3,5,10\}$.
In the context of recommender systems, we are usually interested in recommending top-N items to the user. This is clearly the case for our framework too, items being either employees, apps or website categories. As usual, we consider as relevant items those that are already known in the datasets.

\subsubsection{Cross-Validation}
We use $3$ rounds of \textit{Monte Carlo cross-validation} on the movie production data (due to few non-missing entries) and \textit{3-fold cross-validation} on the OTT and CTR data.
We use the RMSE scores to find the best model via cross-validation.

\subsubsection{Details of Training}
We use a maximum number of $100$ epochs and a tolerance defined by $\frac{\mathcal{L}^{(t)} - \mathcal{L}^{(t-1)}}{\mathcal{L}^{(t)}} < 10^{-6}$ as stopping criteria for training, where $\mathcal{L}^{(t)}$ is the objective cost at epoch $t$.
In all SGD-based algorithms, we use batches of $8$ entries on the VFX data, and of $128$ entries on the OTT and CTR data.
The best number of latent factors $K$ is searched over all possible values in the VFX data, while we use the common values of $K\in\{10,15,20\}$ in the larger OTT and CTR data.
Each matrix factor is initialized with uniformly random numbers in $(0, 1)$, and biases are initialized as zero vectors.

\subsubsection{Learning Algorithms}
The following algorithms are evaluated:
\begin{itemize}
	\setlength{\itemsep}{20pt}
	\setlength{\parskip}{-16pt}
	\setlength{\parsep}{0pt}
	\item \textbf{MF}. Stochastic gradient descent for MF with biases.
	\item \textbf{NMF}. Alternating non-negative least squares for NMF. Each block is updated by projected gradient descent.
	\item \textbf{BMF}. Bounded matrix factorization. To overcome the limitation requiring $x_{max}>1$ in~\cite{kannan2014bounded}, we scale the randomly initialized factors \textbf{W} and \textbf{Z} by $\sqrt{\frac{1}{\max\left(\textbf{W}\cdot\textbf{Z}^T\right)}}$, ensuring feasible bounds at the first step.
	\item \textbf{PMF}. Probabilistic matrix factorization.
	\item \textbf{LMF}. Logistic matrix factorization. Differently from~\cite{johnson2014logistic}, we use explicit data and model each entry in the data matrix to match our framework:
	\begin{equation}
	\begin{aligned}
	\hat{x}_{dn} &=\mathds{P}\left(\text{claims by artist $d$ in discipline $n$ are accepted}\right)\\
	& = \mathds{P}\left(\text{manager of discipline $n$ accepts a claim}\right)\\
	&~~\hspace*{0.3cm}~~~\mathds{P}\left(\text{artist $d$ submits good claims in discipline $n$}\right)\\
	& = \frac{1}{1 + e^{-\gamma_{n}}}\frac{1}{1 + e^{-\beta_{d} - \textbf{w}_d^T\cdot\textbf{z}_n}}.
	\end{aligned}
	\end{equation}
	\item \textbf{EMF}. Expertise matrix factorization. Projections of each row in \textbf{Z} onto the probability simplex are obtained using the $\mathcal{O}\left(K\log K\right)$ efficient algorithm proposed in~\cite{wang2013projection}.
	\item \textbf{SMF}. Survival matrix factorization. $\sigma$ is initialized to $1$.
\end{itemize}

\subsection{Movie Production Data}
Our analysis is driven by data collected at Technicolor, where different disciplines (departments) are responsible for generating VFX in movies (jobs) and employees are referred as artists. 
All data was collected in accordance with appropriate end user agreements and privacy policies.

\begin{figure}[t]
	\setlength\figureheight{3.2cm}
	\setlength\figurewidth{4.12cm}
\begin{tikzpicture}

\definecolor{color0}{rgb}{0.466666666666667,0.666666666666667,0.866666666666667}

\begin{axis}[
height=\figureheight,
tick align=outside,
tick pos=left,
width=\figurewidth,
x grid style={white!69.01960784313725!black},
xlabel style={text width=\figurewidth},
xlabel={Number of disciplines per artist},
xmajorgrids,
xmin=2.45, xmax=7.95,
y grid style={white!69.01960784313725!black},
ylabel={Number of artists},
tick label style={font=\tiny},
ylabel shift={-4pt},
label style={font=\small},
ymajorgrids,
xticklabel style = {rotate=30},
ymin=0, ymax=249.9
]
\draw[fill=color0,draw opacity=0] (axis cs:2.7,0) rectangle (axis cs:3.3,238);
\draw[fill=color0,draw opacity=0] (axis cs:3.7,0) rectangle (axis cs:4.3,60);
\draw[fill=color0,draw opacity=0] (axis cs:4.7,0) rectangle (axis cs:5.3,13);
\draw[fill=color0,draw opacity=0] (axis cs:5.7,0) rectangle (axis cs:6.3,0);
\draw[fill=color0,draw opacity=0] (axis cs:6.7,0) rectangle (axis cs:7.3,1);
%

\path [draw=black, fill opacity=0] (axis cs:2.45,0)
--(axis cs:7.95,0);

\path [draw=black, fill opacity=0] (axis cs:2.45,1)
--(axis cs:7.95,1);

\end{axis}

\end{tikzpicture}	\input{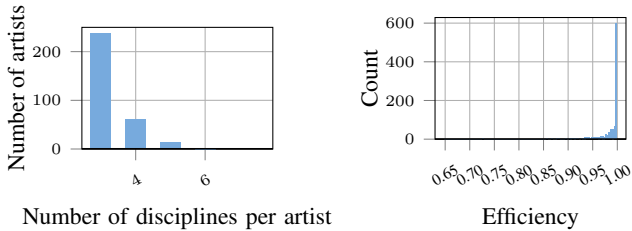}
	\vspace{-0.6cm}
	\caption{Statistics about movie production data.}
	\label{fig:mpc_data}
	\vspace{-0.1cm}
\end{figure}

Our movie production dataset consists of claim records, each with fields: \textit{jobId} (\texttt{int}), \textit{disciplineId} (\texttt{int}), \textit{taskId} (\texttt{int}), \textit{userId} (\texttt{int}), \textit{claimId} (\texttt{int}) and \textit{approved} (\texttt{bool}).
To ensure that each average efficiency is sufficiently representative of the true efficiency of an artist in a discipline, we remove all the entries that result from averaging less than $10$ claims.
This also ensures that, according to Hoeffding's theorem~\cite{hoeffding1963probability}, the approximation introduced in Equation~\eqref{eq:approx_smf} holds true with high probability. 
Moreover, to alleviate the cold-start problem, we remove disciplines that have claims from less than $10$ artists and also drop artists with less than $3$ non-missing entries. 
At the end of these preprocessing steps, we are left with a $312\times 25$ matrix and $1,026$ non-missing entries. Not only is this matrix very sparse ($86.85\%$) but there are also few ratings per user, with most of the artist having worked on three disciplines only. The distribution of efficiencies increases exponentially (Figure~\ref{fig:mpc_data}). 

\begin{table}[t] 
	\centering
	\scriptsize
	\begin{tabular}{ll|cc|cc}
		\hline
		\multirow{3}{*}{\textbf{Method}} & \multirow{3}{*}{\textbf{$K$}} & \multicolumn{2}{c|}{\textbf{RMSE $\left[\times10^{-2}\right]$}} & \multicolumn{2}{c}{\textbf{MAE $\left[\times10^{-2}\right]$}}\\
		& & \multicolumn{2}{c|}{Mean (standard error)} & \multicolumn{2}{c}{Mean (standard error)}\\
		& & \multicolumn{1}{c}{\textit{Training}} & \multicolumn{1}{c|}{\textit{Test}} & \multicolumn{1}{c}{\textit{Training}} & \multicolumn{1}{c}{\textit{Test}} \\ \hline 
		MF &6   &2.50   (0.045)   &2.66  (0.019)   &1.42   (0.021)    &1.79  (0.078)  \\ 
		NMF &1   &2.05   (0.036)   &2.73  (0.016)    &1.19   (0.016)   &1.93  (0.020)  \\ 
		BMF &1   &2.17   (0.069)   &2.64  (0.010)   &1.19   (0.027)    &1.78  (0.008)  \\ 
		PMF &24  &3.53   (0.018)   &3.18  (0.017)   &1.94   (0.010)    &1.99  (0.001)  \\ 
		LMF &21  &5.08   (0.015)   &5.07  (0.043)   &4.42   (0.014)    &4.39  (0.016)  \\  \hline
		EMF &20  &\textbf{0.46 (0.004)}  &2.62 (0.007)   &\textbf{0.36 (0.002)}     &1.60 (0.011) \\ 
		SMF &14  &3.53 (0.079) &\textbf{1.54 (0.017)}  &3.15 (0.026)      &\textbf{1.50 (0.059)}  \\  \hline
	\end{tabular}
	\caption{Prediction errors on the movie production dataset.} 
	\label{tab:mpc_results}
\end{table}

Table~\ref{tab:mpc_results} reports prediction errors of each algorithm on this dataset. Precision and recall values are listed in Table~\ref{tab:mpc_tab}.

\subsection{Over-The-Top Data}
\begin{figure}[t]
	\setlength\figureheight{3.2cm}
	\setlength\figurewidth{4.12cm}
\begin{tikzpicture}

\definecolor{color0}{rgb}{0.466666666666667,0.666666666666667,0.866666666666667}

\begin{axis}[
height=\figureheight,
tick align=outside,
tick pos=left,
width=\figurewidth,
x grid style={white!69.01960784313725!black},
xlabel={Number of apps per user},
xmajorgrids,
xmin=8.4, xmax=37,
y grid style={white!69.01960784313725!black},
ylabel={Number of users},
tick label style={font=\tiny},
ylabel shift={-4pt},
label style={font=\small},
ymajorgrids,
ymin=0, ymax=271.95
]
\draw[fill=color0,draw opacity=0] (axis cs:9.7,0) rectangle (axis cs:10.3,259);
\draw[fill=color0,draw opacity=0] (axis cs:10.7,0) rectangle (axis cs:11.3,191);
\draw[fill=color0,draw opacity=0] (axis cs:11.7,0) rectangle (axis cs:12.3,134);
\draw[fill=color0,draw opacity=0] (axis cs:12.7,0) rectangle (axis cs:13.3,81);
\draw[fill=color0,draw opacity=0] (axis cs:13.7,0) rectangle (axis cs:14.3,73);
\draw[fill=color0,draw opacity=0] (axis cs:14.7,0) rectangle (axis cs:15.3,55);
\draw[fill=color0,draw opacity=0] (axis cs:15.7,0) rectangle (axis cs:16.3,27);
\draw[fill=color0,draw opacity=0] (axis cs:16.7,0) rectangle (axis cs:17.3,24);
\draw[fill=color0,draw opacity=0] (axis cs:17.7,0) rectangle (axis cs:18.3,23);
\draw[fill=color0,draw opacity=0] (axis cs:18.7,0) rectangle (axis cs:19.3,19);
\draw[fill=color0,draw opacity=0] (axis cs:19.7,0) rectangle (axis cs:20.3,10);
\draw[fill=color0,draw opacity=0] (axis cs:20.7,0) rectangle (axis cs:21.3,2);
\draw[fill=color0,draw opacity=0] (axis cs:21.7,0) rectangle (axis cs:22.3,5);
\draw[fill=color0,draw opacity=0] (axis cs:22.7,0) rectangle (axis cs:23.3,10);
\draw[fill=color0,draw opacity=0] (axis cs:23.7,0) rectangle (axis cs:24.3,2);
\draw[fill=color0,draw opacity=0] (axis cs:24.7,0) rectangle (axis cs:25.3,4);
\draw[fill=color0,draw opacity=0] (axis cs:25.7,0) rectangle (axis cs:26.3,3);
\draw[fill=color0,draw opacity=0] (axis cs:26.7,0) rectangle (axis cs:27.3,2);
\draw[fill=color0,draw opacity=0] (axis cs:27.7,0) rectangle (axis cs:28.3,2);
\draw[fill=color0,draw opacity=0] (axis cs:28.7,0) rectangle (axis cs:29.3,2);
\draw[fill=color0,draw opacity=0] (axis cs:29.7,0) rectangle (axis cs:30.3,3);
\draw[fill=color0,draw opacity=0] (axis cs:30.7,0) rectangle (axis cs:31.3,2);
\draw[fill=color0,draw opacity=0] (axis cs:31.7,0) rectangle (axis cs:32.3,0);
\draw[fill=color0,draw opacity=0] (axis cs:32.7,0) rectangle (axis cs:33.3,0);
\draw[fill=color0,draw opacity=0] (axis cs:33.7,0) rectangle (axis cs:34.3,0);
\draw[fill=color0,draw opacity=0] (axis cs:34.7,0) rectangle (axis cs:35.3,1);
%

\path [draw=black, fill opacity=0] (axis cs:8.4,0)
--(axis cs:37,0);

\path [draw=black, fill opacity=0] (axis cs:8.4,1)
--(axis cs:37,1);

\end{axis}

\end{tikzpicture}	\input{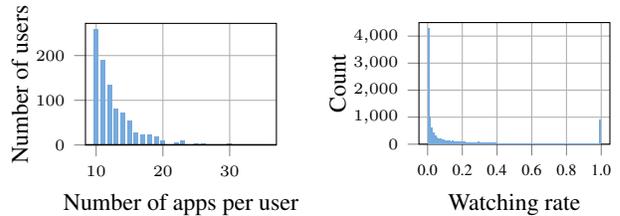}
	\vspace{-0.08cm}
	\caption{Statistics about over-the-top data.}
	\label{fig:ott_data}
\end{figure}

Our second application consists of apps usage on over-the-top devices where the goal is to recommend apps to customers.
Our OTT dataset consists of number of views of a given app by any user. This data is mapped to users' watching rates $\in [0,1]$ by dividing by the user's maximum number of views.
Similarly to what we did for the VFX data, we remove apps that have less than $15$ viewers and users who watched less than $10$ apps.
After preprocessing, we are left with $934$ users and $140$ apps and a matrix that is extremely sparse ($99.91\%$). The distribution of watching rates is shown in Figure~\ref{fig:ott_data}.

Tables~\ref{tab:ott_15_tab} and \ref{tab:ott_20_tab} show precision and recall values for $K$ equal to $15$ and $20$, respectively. Due to space constraints, results for $K = 10$ are omitted but followed similar patterns.

\subsection{Click-Through Rate Data}
In our third application, we consider the task of placing online advertisements in categories of websites (e.g., Entertainment, Finance, etc.) to maximize their click-through rate.

\begin{figure}
	\setlength\figureheight{3.2cm}
	\setlength\figurewidth{4.12cm}
\begin{tikzpicture}

\definecolor{color0}{rgb}{0.466666666666667,0.666666666666667,0.866666666666667}

\begin{axis}[
height=\figureheight,
tick align=outside,
tick pos=left,
width=\figurewidth,
x grid style={white!69.01960784313725!black},
xlabel={Number of categories per ad},
xmajorgrids,
xmin=0.7, xmax=88.7,
y grid style={white!69.01960784313725!black},
ylabel={Number of ads},
tick label style={font=\tiny},
ylabel shift={-4pt},
label style={font=\small},
ymajorgrids,
ymin=0, ymax=1907.85
]
\draw[fill=color0,draw opacity=0] (axis cs:4.7,0) rectangle (axis cs:5.3,1817);
\draw[fill=color0,draw opacity=0] (axis cs:5.7,0) rectangle (axis cs:6.3,1438);
\draw[fill=color0,draw opacity=0] (axis cs:6.7,0) rectangle (axis cs:7.3,1202);
\draw[fill=color0,draw opacity=0] (axis cs:7.7,0) rectangle (axis cs:8.3,982);
\draw[fill=color0,draw opacity=0] (axis cs:8.7,0) rectangle (axis cs:9.3,868);
\draw[fill=color0,draw opacity=0] (axis cs:9.7,0) rectangle (axis cs:10.3,719);
\draw[fill=color0,draw opacity=0] (axis cs:10.7,0) rectangle (axis cs:11.3,684);
\draw[fill=color0,draw opacity=0] (axis cs:11.7,0) rectangle (axis cs:12.3,596);
\draw[fill=color0,draw opacity=0] (axis cs:12.7,0) rectangle (axis cs:13.3,561);
\draw[fill=color0,draw opacity=0] (axis cs:13.7,0) rectangle (axis cs:14.3,455);
\draw[fill=color0,draw opacity=0] (axis cs:14.7,0) rectangle (axis cs:15.3,454);
\draw[fill=color0,draw opacity=0] (axis cs:15.7,0) rectangle (axis cs:16.3,363);
\draw[fill=color0,draw opacity=0] (axis cs:16.7,0) rectangle (axis cs:17.3,345);
\draw[fill=color0,draw opacity=0] (axis cs:17.7,0) rectangle (axis cs:18.3,337);
\draw[fill=color0,draw opacity=0] (axis cs:18.7,0) rectangle (axis cs:19.3,295);
\draw[fill=color0,draw opacity=0] (axis cs:19.7,0) rectangle (axis cs:20.3,283);
\draw[fill=color0,draw opacity=0] (axis cs:20.7,0) rectangle (axis cs:21.3,247);
\draw[fill=color0,draw opacity=0] (axis cs:21.7,0) rectangle (axis cs:22.3,262);
\draw[fill=color0,draw opacity=0] (axis cs:22.7,0) rectangle (axis cs:23.3,228);
\draw[fill=color0,draw opacity=0] (axis cs:23.7,0) rectangle (axis cs:24.3,189);
\draw[fill=color0,draw opacity=0] (axis cs:24.7,0) rectangle (axis cs:25.3,201);
\draw[fill=color0,draw opacity=0] (axis cs:25.7,0) rectangle (axis cs:26.3,175);
\draw[fill=color0,draw opacity=0] (axis cs:26.7,0) rectangle (axis cs:27.3,167);
\draw[fill=color0,draw opacity=0] (axis cs:27.7,0) rectangle (axis cs:28.3,149);
\draw[fill=color0,draw opacity=0] (axis cs:28.7,0) rectangle (axis cs:29.3,138);
\draw[fill=color0,draw opacity=0] (axis cs:29.7,0) rectangle (axis cs:30.3,145);
\draw[fill=color0,draw opacity=0] (axis cs:30.7,0) rectangle (axis cs:31.3,112);
\draw[fill=color0,draw opacity=0] (axis cs:31.7,0) rectangle (axis cs:32.3,143);
\draw[fill=color0,draw opacity=0] (axis cs:32.7,0) rectangle (axis cs:33.3,108);
\draw[fill=color0,draw opacity=0] (axis cs:33.7,0) rectangle (axis cs:34.3,112);
\draw[fill=color0,draw opacity=0] (axis cs:34.7,0) rectangle (axis cs:35.3,105);
\draw[fill=color0,draw opacity=0] (axis cs:35.7,0) rectangle (axis cs:36.3,102);
\draw[fill=color0,draw opacity=0] (axis cs:36.7,0) rectangle (axis cs:37.3,86);
\draw[fill=color0,draw opacity=0] (axis cs:37.7,0) rectangle (axis cs:38.3,98);
\draw[fill=color0,draw opacity=0] (axis cs:38.7,0) rectangle (axis cs:39.3,94);
\draw[fill=color0,draw opacity=0] (axis cs:39.7,0) rectangle (axis cs:40.3,108);
\draw[fill=color0,draw opacity=0] (axis cs:40.7,0) rectangle (axis cs:41.3,65);
\draw[fill=color0,draw opacity=0] (axis cs:41.7,0) rectangle (axis cs:42.3,84);
\draw[fill=color0,draw opacity=0] (axis cs:42.7,0) rectangle (axis cs:43.3,73);
\draw[fill=color0,draw opacity=0] (axis cs:43.7,0) rectangle (axis cs:44.3,64);
\draw[fill=color0,draw opacity=0] (axis cs:44.7,0) rectangle (axis cs:45.3,50);
\draw[fill=color0,draw opacity=0] (axis cs:45.7,0) rectangle (axis cs:46.3,46);
\draw[fill=color0,draw opacity=0] (axis cs:46.7,0) rectangle (axis cs:47.3,62);
\draw[fill=color0,draw opacity=0] (axis cs:47.7,0) rectangle (axis cs:48.3,54);
\draw[fill=color0,draw opacity=0] (axis cs:48.7,0) rectangle (axis cs:49.3,54);
\draw[fill=color0,draw opacity=0] (axis cs:49.7,0) rectangle (axis cs:50.3,56);
\draw[fill=color0,draw opacity=0] (axis cs:50.7,0) rectangle (axis cs:51.3,46);
\draw[fill=color0,draw opacity=0] (axis cs:51.7,0) rectangle (axis cs:52.3,47);
\draw[fill=color0,draw opacity=0] (axis cs:52.7,0) rectangle (axis cs:53.3,42);
\draw[fill=color0,draw opacity=0] (axis cs:53.7,0) rectangle (axis cs:54.3,32);
\draw[fill=color0,draw opacity=0] (axis cs:54.7,0) rectangle (axis cs:55.3,34);
\draw[fill=color0,draw opacity=0] (axis cs:55.7,0) rectangle (axis cs:56.3,46);
\draw[fill=color0,draw opacity=0] (axis cs:56.7,0) rectangle (axis cs:57.3,42);
\draw[fill=color0,draw opacity=0] (axis cs:57.7,0) rectangle (axis cs:58.3,40);
\draw[fill=color0,draw opacity=0] (axis cs:58.7,0) rectangle (axis cs:59.3,28);
\draw[fill=color0,draw opacity=0] (axis cs:59.7,0) rectangle (axis cs:60.3,26);
\draw[fill=color0,draw opacity=0] (axis cs:60.7,0) rectangle (axis cs:61.3,30);
\draw[fill=color0,draw opacity=0] (axis cs:61.7,0) rectangle (axis cs:62.3,22);
\draw[fill=color0,draw opacity=0] (axis cs:62.7,0) rectangle (axis cs:63.3,32);
\draw[fill=color0,draw opacity=0] (axis cs:63.7,0) rectangle (axis cs:64.3,28);
\draw[fill=color0,draw opacity=0] (axis cs:64.7,0) rectangle (axis cs:65.3,22);
\draw[fill=color0,draw opacity=0] (axis cs:65.7,0) rectangle (axis cs:66.3,22);
\draw[fill=color0,draw opacity=0] (axis cs:66.7,0) rectangle (axis cs:67.3,20);
\draw[fill=color0,draw opacity=0] (axis cs:67.7,0) rectangle (axis cs:68.3,15);
\draw[fill=color0,draw opacity=0] (axis cs:68.7,0) rectangle (axis cs:69.3,7);
\draw[fill=color0,draw opacity=0] (axis cs:69.7,0) rectangle (axis cs:70.3,10);
\draw[fill=color0,draw opacity=0] (axis cs:70.7,0) rectangle (axis cs:71.3,14);
\draw[fill=color0,draw opacity=0] (axis cs:71.7,0) rectangle (axis cs:72.3,12);
\draw[fill=color0,draw opacity=0] (axis cs:72.7,0) rectangle (axis cs:73.3,11);
\draw[fill=color0,draw opacity=0] (axis cs:73.7,0) rectangle (axis cs:74.3,10);
\draw[fill=color0,draw opacity=0] (axis cs:74.7,0) rectangle (axis cs:75.3,5);
\draw[fill=color0,draw opacity=0] (axis cs:75.7,0) rectangle (axis cs:76.3,7);
\draw[fill=color0,draw opacity=0] (axis cs:76.7,0) rectangle (axis cs:77.3,8);
\draw[fill=color0,draw opacity=0] (axis cs:77.7,0) rectangle (axis cs:78.3,4);
\draw[fill=color0,draw opacity=0] (axis cs:78.7,0) rectangle (axis cs:79.3,3);
\draw[fill=color0,draw opacity=0] (axis cs:79.7,0) rectangle (axis cs:80.3,3);
\draw[fill=color0,draw opacity=0] (axis cs:80.7,0) rectangle (axis cs:81.3,1);
\draw[fill=color0,draw opacity=0] (axis cs:81.7,0) rectangle (axis cs:82.3,0);
\draw[fill=color0,draw opacity=0] (axis cs:82.7,0) rectangle (axis cs:83.3,1);
\draw[fill=color0,draw opacity=0] (axis cs:83.7,0) rectangle (axis cs:84.3,1);
%

\path [draw=black, fill opacity=0] (axis cs:0.7,0)
--(axis cs:88.7,0);

\path [draw=black, fill opacity=0] (axis cs:0.7,1)
--(axis cs:88.7,1);

\end{axis}

\end{tikzpicture}	
\begin{tikzpicture}

\definecolor{color0}{rgb}{0.466666666666667,0.666666666666667,0.866666666666667}

\begin{axis}[
height=\figureheight,
tick align=outside,
tick pos=left,
width=\figurewidth,
x grid style={white!69.01960784313725!black},
xlabel={Click-through rate},
xmajorgrids,
xmin=-0.05, xmax=1.05,
xtick={-0.2,0,0.2,0.4,0.6,0.8,1,1.2},
xticklabels={−0.2,0.0,0.2,0.4,0.6,0.8,1.0,1.2},
y grid style={white!69.01960784313725!black},
ylabel={Count},
tick label style={font=\tiny},
ylabel shift={-4pt},
label style={font=\small},
ymajorgrids,
ymin=0, ymax=29109.15
]
\draw[fill=color0,draw opacity=0] (axis cs:0,0) rectangle (axis cs:0.01,27723);
\draw[fill=color0,draw opacity=0] (axis cs:0.01,0) rectangle (axis cs:0.02,2359);
\draw[fill=color0,draw opacity=0] (axis cs:0.02,0) rectangle (axis cs:0.03,4970);
\draw[fill=color0,draw opacity=0] (axis cs:0.03,0) rectangle (axis cs:0.04,6355);
\draw[fill=color0,draw opacity=0] (axis cs:0.04,0) rectangle (axis cs:0.05,8020);
\draw[fill=color0,draw opacity=0] (axis cs:0.05,0) rectangle (axis cs:0.06,9947);
\draw[fill=color0,draw opacity=0] (axis cs:0.06,0) rectangle (axis cs:0.07,9336);
\draw[fill=color0,draw opacity=0] (axis cs:0.07,0) rectangle (axis cs:0.08,10196);
\draw[fill=color0,draw opacity=0] (axis cs:0.08,0) rectangle (axis cs:0.09,9444);
\draw[fill=color0,draw opacity=0] (axis cs:0.09,0) rectangle (axis cs:0.1,9589);
\draw[fill=color0,draw opacity=0] (axis cs:0.1,0) rectangle (axis cs:0.11,11341);
\draw[fill=color0,draw opacity=0] (axis cs:0.11,0) rectangle (axis cs:0.12,7427);
\draw[fill=color0,draw opacity=0] (axis cs:0.12,0) rectangle (axis cs:0.13,6899);
\draw[fill=color0,draw opacity=0] (axis cs:0.13,0) rectangle (axis cs:0.14,7117);
\draw[fill=color0,draw opacity=0] (axis cs:0.14,0) rectangle (axis cs:0.15,6919);
\draw[fill=color0,draw opacity=0] (axis cs:0.15,0) rectangle (axis cs:0.16,7762);
\draw[fill=color0,draw opacity=0] (axis cs:0.16,0) rectangle (axis cs:0.17,7766);
\draw[fill=color0,draw opacity=0] (axis cs:0.17,0) rectangle (axis cs:0.18,5512);
\draw[fill=color0,draw opacity=0] (axis cs:0.18,0) rectangle (axis cs:0.19,7744);
\draw[fill=color0,draw opacity=0] (axis cs:0.19,0) rectangle (axis cs:0.2,3703);
\draw[fill=color0,draw opacity=0] (axis cs:0.2,0) rectangle (axis cs:0.21,8584);
\draw[fill=color0,draw opacity=0] (axis cs:0.21,0) rectangle (axis cs:0.22,5501);
\draw[fill=color0,draw opacity=0] (axis cs:0.22,0) rectangle (axis cs:0.23,4023);
\draw[fill=color0,draw opacity=0] (axis cs:0.23,0) rectangle (axis cs:0.24,5222);
\draw[fill=color0,draw opacity=0] (axis cs:0.24,0) rectangle (axis cs:0.25,2519);
\draw[fill=color0,draw opacity=0] (axis cs:0.25,0) rectangle (axis cs:0.26,5915);
\draw[fill=color0,draw opacity=0] (axis cs:0.26,0) rectangle (axis cs:0.27,3834);
\draw[fill=color0,draw opacity=0] (axis cs:0.27,0) rectangle (axis cs:0.28,4348);
\draw[fill=color0,draw opacity=0] (axis cs:0.28,0) rectangle (axis cs:0.29,3329);
\draw[fill=color0,draw opacity=0] (axis cs:0.29,0) rectangle (axis cs:0.3,2544);
\draw[fill=color0,draw opacity=0] (axis cs:0.3,0) rectangle (axis cs:0.31,5125);
\draw[fill=color0,draw opacity=0] (axis cs:0.31,0) rectangle (axis cs:0.32,2668);
\draw[fill=color0,draw opacity=0] (axis cs:0.32,0) rectangle (axis cs:0.33,1813);
\draw[fill=color0,draw opacity=0] (axis cs:0.33,0) rectangle (axis cs:0.34,3710);
\draw[fill=color0,draw opacity=0] (axis cs:0.34,0) rectangle (axis cs:0.35,1954);
\draw[fill=color0,draw opacity=0] (axis cs:0.35,0) rectangle (axis cs:0.36,2152);
\draw[fill=color0,draw opacity=0] (axis cs:0.36,0) rectangle (axis cs:0.37,2682);
\draw[fill=color0,draw opacity=0] (axis cs:0.37,0) rectangle (axis cs:0.38,1632);
\draw[fill=color0,draw opacity=0] (axis cs:0.38,0) rectangle (axis cs:0.39,2173);
\draw[fill=color0,draw opacity=0] (axis cs:0.39,0) rectangle (axis cs:0.4,980);
\draw[fill=color0,draw opacity=0] (axis cs:0.4,0) rectangle (axis cs:0.41,2802);
\draw[fill=color0,draw opacity=0] (axis cs:0.41,0) rectangle (axis cs:0.42,1814);
\draw[fill=color0,draw opacity=0] (axis cs:0.42,0) rectangle (axis cs:0.43,1577);
\draw[fill=color0,draw opacity=0] (axis cs:0.43,0) rectangle (axis cs:0.44,1026);
\draw[fill=color0,draw opacity=0] (axis cs:0.44,0) rectangle (axis cs:0.45,987);
\draw[fill=color0,draw opacity=0] (axis cs:0.45,0) rectangle (axis cs:0.46,1593);
\draw[fill=color0,draw opacity=0] (axis cs:0.46,0) rectangle (axis cs:0.47,1343);
\draw[fill=color0,draw opacity=0] (axis cs:0.47,0) rectangle (axis cs:0.48,984);
\draw[fill=color0,draw opacity=0] (axis cs:0.48,0) rectangle (axis cs:0.49,657);
\draw[fill=color0,draw opacity=0] (axis cs:0.49,0) rectangle (axis cs:0.5,303);
\draw[fill=color0,draw opacity=0] (axis cs:0.5,0) rectangle (axis cs:0.51,2228);
\draw[fill=color0,draw opacity=0] (axis cs:0.51,0) rectangle (axis cs:0.52,488);
\draw[fill=color0,draw opacity=0] (axis cs:0.52,0) rectangle (axis cs:0.53,735);
\draw[fill=color0,draw opacity=0] (axis cs:0.53,0) rectangle (axis cs:0.54,834);
\draw[fill=color0,draw opacity=0] (axis cs:0.54,0) rectangle (axis cs:0.55,771);
\draw[fill=color0,draw opacity=0] (axis cs:0.55,0) rectangle (axis cs:0.56,500);
\draw[fill=color0,draw opacity=0] (axis cs:0.56,0) rectangle (axis cs:0.57,462);
\draw[fill=color0,draw opacity=0] (axis cs:0.57,0) rectangle (axis cs:0.58,545);
\draw[fill=color0,draw opacity=0] (axis cs:0.58,0) rectangle (axis cs:0.59,569);
\draw[fill=color0,draw opacity=0] (axis cs:0.59,0) rectangle (axis cs:0.6,260);
\draw[fill=color0,draw opacity=0] (axis cs:0.6,0) rectangle (axis cs:0.61,802);
\draw[fill=color0,draw opacity=0] (axis cs:0.61,0) rectangle (axis cs:0.62,442);
\draw[fill=color0,draw opacity=0] (axis cs:0.62,0) rectangle (axis cs:0.63,290);
\draw[fill=color0,draw opacity=0] (axis cs:0.63,0) rectangle (axis cs:0.64,439);
\draw[fill=color0,draw opacity=0] (axis cs:0.64,0) rectangle (axis cs:0.65,284);
\draw[fill=color0,draw opacity=0] (axis cs:0.65,0) rectangle (axis cs:0.66,203);
\draw[fill=color0,draw opacity=0] (axis cs:0.66,0) rectangle (axis cs:0.67,334);
\draw[fill=color0,draw opacity=0] (axis cs:0.67,0) rectangle (axis cs:0.68,106);
\draw[fill=color0,draw opacity=0] (axis cs:0.68,0) rectangle (axis cs:0.69,206);
\draw[fill=color0,draw opacity=0] (axis cs:0.69,0) rectangle (axis cs:0.7,351);
\draw[fill=color0,draw opacity=0] (axis cs:0.7,0) rectangle (axis cs:0.71,94);
\draw[fill=color0,draw opacity=0] (axis cs:0.71,0) rectangle (axis cs:0.72,129);
\draw[fill=color0,draw opacity=0] (axis cs:0.72,0) rectangle (axis cs:0.73,177);
\draw[fill=color0,draw opacity=0] (axis cs:0.73,0) rectangle (axis cs:0.74,103);
\draw[fill=color0,draw opacity=0] (axis cs:0.74,0) rectangle (axis cs:0.75,43);
\draw[fill=color0,draw opacity=0] (axis cs:0.75,0) rectangle (axis cs:0.76,141);
\draw[fill=color0,draw opacity=0] (axis cs:0.76,0) rectangle (axis cs:0.77,98);
\draw[fill=color0,draw opacity=0] (axis cs:0.77,0) rectangle (axis cs:0.78,50);
\draw[fill=color0,draw opacity=0] (axis cs:0.78,0) rectangle (axis cs:0.79,62);
\draw[fill=color0,draw opacity=0] (axis cs:0.79,0) rectangle (axis cs:0.8,20);
\draw[fill=color0,draw opacity=0] (axis cs:0.8,0) rectangle (axis cs:0.81,108);
\draw[fill=color0,draw opacity=0] (axis cs:0.81,0) rectangle (axis cs:0.82,64);
\draw[fill=color0,draw opacity=0] (axis cs:0.82,0) rectangle (axis cs:0.83,27);
\draw[fill=color0,draw opacity=0] (axis cs:0.83,0) rectangle (axis cs:0.84,43);
\draw[fill=color0,draw opacity=0] (axis cs:0.84,0) rectangle (axis cs:0.85,25);
\draw[fill=color0,draw opacity=0] (axis cs:0.85,0) rectangle (axis cs:0.86,23);
\draw[fill=color0,draw opacity=0] (axis cs:0.86,0) rectangle (axis cs:0.87,17);
\draw[fill=color0,draw opacity=0] (axis cs:0.87,0) rectangle (axis cs:0.88,12);
\draw[fill=color0,draw opacity=0] (axis cs:0.88,0) rectangle (axis cs:0.89,13);
\draw[fill=color0,draw opacity=0] (axis cs:0.89,0) rectangle (axis cs:0.9,4);
\draw[fill=color0,draw opacity=0] (axis cs:0.9,0) rectangle (axis cs:0.91,37);
\draw[fill=color0,draw opacity=0] (axis cs:0.91,0) rectangle (axis cs:0.92,11);
\draw[fill=color0,draw opacity=0] (axis cs:0.92,0) rectangle (axis cs:0.93,10);
\draw[fill=color0,draw opacity=0] (axis cs:0.93,0) rectangle (axis cs:0.94,5);
\draw[fill=color0,draw opacity=0] (axis cs:0.94,0) rectangle (axis cs:0.95,5);
\draw[fill=color0,draw opacity=0] (axis cs:0.95,0) rectangle (axis cs:0.96,2);
\draw[fill=color0,draw opacity=0] (axis cs:0.96,0) rectangle (axis cs:0.97,1);
\draw[fill=color0,draw opacity=0] (axis cs:0.97,0) rectangle (axis cs:0.98,1);
\draw[fill=color0,draw opacity=0] (axis cs:0.98,0) rectangle (axis cs:0.99,0);
\draw[fill=color0,draw opacity=0] (axis cs:0.99,0) rectangle (axis cs:1,9);
%

\path [draw=black, fill opacity=0] (axis cs:-0.05,0)
--(axis cs:1.05,0);

\path [draw=black, fill opacity=0] (axis cs:-0.05,1)
--(axis cs:1.05,1);

\end{axis}

\end{tikzpicture}
	\vspace{-0.2cm}
	\caption{Statistics about click-through rate data.}
	\label{fig:ctr_data}
\end{figure}
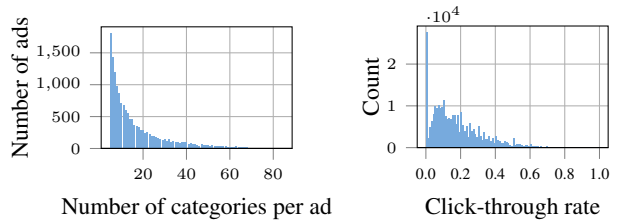

To simulate this scenario, we use publicly available data from a Kaggle competition~\footnote{URL: \url{https://www.kaggle.com/c/outbrain-click-prediction}.} run by the advertisement company Outbrain.
This dataset contains users' webpage views and clicks on multiple publisher sites in the United States in a two-week period in June 2016.
By using additional semantic attributes of the webpages, we can formulate the problem of maximizing the CTR of an advertisement by finding the best categories to display it.
After removing advertisements displayed in less than $5$ categories and categories with less than $10$ advertisements, we are left with $15,647$ advertisements and $85$ webpage categories, resulting in a $79.99\%$-sparse matrix and a distribution of CTR skewed to the right (Figure~\ref{fig:ctr_data}).

Tables~\ref{tab:ctr_10_tab}, \ref{tab:ctr_15_tab} and \ref{tab:ctr_20_tab} show precision and recall values for number of latent factors $K$ equal to $10$, $15$ and $20$, respectively. 

\subsection{Discussion}
In our VFX data, SMF achieves the lowest prediction errors and EMF outperforms every other method in precision and recall with values three times higher than the second best model. Here, the hypotheses of EMF closely match artists in our VFX framework, while they do not entirely hold for ads in CTR and users in OTT.
One of the strongest points of PMF is its ability to generalize considerably well for users with very few ratings. However, here -- where the maximum number of entries per user is just $7$ -- we see that EMF outperforms PMF in both prediction accuracy and quality of recommendations.
In Table~\ref{tab:mpc_results}, we also notice an interesting empirical finding: constrained models are capable of better exploiting the underlying structure relying on a higher latent dimensionality.

Results in OTT and CTR datasets instead show how SMF is a very general model and outperforms every other algorithm in terms of precision and recall, regardless of the chosen number of latent factors $K$. In OTT, the gap is wider for smaller values of $K$; in CTR, EMF consistently achieves the second highest precision.
We believe that SMF underperforms in the VFX data due to the very few available entries rather than its sparsity (OTT is much sparser). Here, SMF's performance is comparable to other bounded models such as BMF and LMF.

\section{Conclusion}\label{sec:conclusion}
In this paper, we introduced \textit{EMF} and \textit{SMF}: two novel structured MF methods to predict entries in the unit interval.
We showed that these models provide better recommendations than popular MF techniques on three real-world datasets.
Hence, it is beneficial to explicitly model entries in the unit interval, motivating further research in this domain.

In addition, SMF is more than just one method: it is the simplest in a class of probabilistic models. In this paper, we assumed users' quality of work to be normally distributed and managers to be modeled by a scalar threshold variable. In future work, we plan to extend this model to include more complex assumptions depending on the application domains.

\clearpage

\begin{table*}[t] 
	\centering
	\scriptsize
	\begin{tabular}{l||c|c|c|c||c|c|c|c}
		\hline
		\multirow{2}{*}{\textbf{Method}} &
		\multicolumn{4}{c||}{\textbf{Precision}} & \multicolumn{4}{c}{\textbf{Recall}}\\
		& \multicolumn{1}{c|}{@2 $\left[\times10^{-2}\right]$} & \multicolumn{1}{c|}{@3 $\left[\times10^{-2}\right]$}& \multicolumn{1}{c|}{@5 $\left[\times10^{-2}\right]$}& \multicolumn{1}{c||}{@10 $\left[\times10^{-2}\right]$}&
		\multicolumn{1}{c|}{@2 $\left[\times10^{-2}\right]$} & \multicolumn{1}{c|}{@3 $\left[\times10^{-2}\right]$}& \multicolumn{1}{c|}{@5 $\left[\times10^{-2}\right]$}& \multicolumn{1}{c}{@10 $\left[\times10^{-2}\right]$}\\
		\hline 
		MF &2.67 (0.544) &3.11 (0.363) &4.80 (0.377) &5.47 (0.393) &0.19 (0.065) &0.37 (0.037) &1.01 (0.042) &2.27 (0.268) \\
		NMF &1.33 (0.544) &2.22 (0.363) &3.47 (0.435) &4.67 (0.435) &0.10 (0.042) &0.57 (0.190) &1.42 (0.265) &2.67 (0.243) \\
		BMF &0.00 (0.000) &0.89 (0.726) &2.13 (0.576) &2.67 (0.109) &0.00 (0.000) &0.05 (0.037) &0.53 (0.038) &1.40 (0.010) \\
		PMF &1.33 (0.544) &3.11 (0.363) &3.47 (0.218) &3.20 (0.189) &0.13 (0.059) &0.37 (0.026) &0.76 (0.116) &1.52 (0.021) \\
		LMF &1.33 (0.544) &0.89 (0.363) &2.67 (0.576) &4.67 (0.288) &0.24 (0.099) &0.24 (0.099) &0.88 (0.044) &2.95 (0.226) \\ \hline
		EMF &\textbf{4.67 (0.544)} &\textbf{6.22 (0.363)} &\textbf{10.40 (0.998)} &\textbf{19.07 (1.039)} &\textbf{0.52 (0.125)} &\textbf{1.14 (0.091)} &\textbf{2.34 (0.132)} &\textbf{8.39 (0.636)} \\
		SMF &4.00 (0.000) &2.67 (0.000) &2.13 (0.218) &2.93 (0.288) &0.29 (0.009) &0.29 (0.009) &0.53 (0.093) &2.82 (0.204) \\ \hline
	\end{tabular}
	\caption{Precision and Recall (mean and standard error) on the movie production dataset.} 
	\label{tab:mpc_tab}
\end{table*}

\begin{table*}[t] 
	\centering
	\scriptsize
	\begin{tabular}{l||c|c|c|c||c|c|c|c}
		\hline
		\multirow{2}{*}{\textbf{Method}} &
		\multicolumn{4}{c||}{\textbf{Precision}} & \multicolumn{4}{c}{\textbf{Recall}}\\
		& \multicolumn{1}{c|}{@2 $\left[\times10^{-2}\right]$} & \multicolumn{1}{c|}{@3 $\left[\times10^{-2}\right]$}& \multicolumn{1}{c|}{@5 $\left[\times10^{-2}\right]$}& \multicolumn{1}{c||}{@10 $\left[\times10^{-2}\right]$}&
		\multicolumn{1}{c|}{@2 $\left[\times10^{-2}\right]$} & \multicolumn{1}{c|}{@3 $\left[\times10^{-2}\right]$}& \multicolumn{1}{c|}{@5 $\left[\times10^{-2}\right]$}& \multicolumn{1}{c}{@10 $\left[\times10^{-2}\right]$}\\
		\hline 
		
		MF &2.02 (0.194) &2.22 (0.130) &2.48 (0.216) &3.17 (0.103) &0.20 (0.049) &0.29 (0.044) &0.46 (0.055) &1.17 (0.118) \\
		NMF &2.74 (0.257) &2.78 (0.467) &3.71 (0.294) &3.21 (0.202) &0.24 (0.024) &0.35 (0.055) &0.70 (0.034) &1.17 (0.074) \\
		BMF &0.95 (0.097) &1.75 (0.130) &2.67 (0.318) &3.24 (0.395) &0.03 (0.010) &0.11 (0.011) &0.27 (0.010) &0.69 (0.067) \\
		PMF &0.24 (0.097) &0.24 (0.112) &0.52 (0.103) &1.48 (0.103) &0.03 (0.021) &0.03 (0.019) &0.09 (0.023) &0.47 (0.016) \\
		LMF &0.24 (0.097) &0.48 (0.194) &0.52 (0.039) &0.90 (0.136) &0.00 (0.003) &0.04 (0.017) &0.08 (0.013) &0.35 (0.073) \\ \hline
		EMF &0.71 (0.000) &1.03 (0.171) &2.00 (0.178) &2.81 (0.248) &0.08 (0.021) &0.12 (0.009) &0.26 (0.054) &0.64 (0.064) \\
		SMF &\textbf{32.62 (0.350)} &\textbf{30.71 (0.194)} &\textbf{25.48 (0.433)} &\textbf{18.60 (0.402)} &\textbf{2.69 (0.063)} &\textbf{3.40 (0.087)} &\textbf{4.18 (0.109)} &\textbf{5.70 (0.130)} \\ \hline
	\end{tabular}
	\caption{Precision and Recall (mean and standard error) on the OTT dataset for $K = 15$.} 
	\label{tab:ott_15_tab}
\end{table*}
\begin{table*}[t] 
	\centering
	\scriptsize
	\begin{tabular}{l||c|c|c|c||c|c|c|c}
		\hline
		\multirow{2}{*}{\textbf{Method}} &
		\multicolumn{4}{c||}{\textbf{Precision}} & \multicolumn{4}{c}{\textbf{Recall}}\\
		& \multicolumn{1}{c|}{@2 $\left[\times10^{-2}\right]$} & \multicolumn{1}{c|}{@3 $\left[\times10^{-2}\right]$}& \multicolumn{1}{c|}{@5 $\left[\times10^{-2}\right]$}& \multicolumn{1}{c||}{@10 $\left[\times10^{-2}\right]$}&
		\multicolumn{1}{c|}{@2 $\left[\times10^{-2}\right]$} & \multicolumn{1}{c|}{@3 $\left[\times10^{-2}\right]$}& \multicolumn{1}{c|}{@5 $\left[\times10^{-2}\right]$}& \multicolumn{1}{c}{@10 $\left[\times10^{-2}\right]$}\\
		\hline 
		MF &2.02 (0.424) &2.06 (0.282) &3.00 (0.135) &3.43 (0.221) &0.15 (0.035) &0.22 (0.033) &0.59 (0.067) &1.34 (0.165) \\
		NMF &1.07 (0.292) &1.51 (0.259) &1.43 (0.178) &1.45 (0.237) &0.11 (0.034) &0.22 (0.038) &0.35 (0.047) &0.67 (0.121) \\
		BMF &1.90 (0.097) &2.70 (0.282) &4.14 (0.486) &4.26 (0.051) &0.09 (0.029) &0.21 (0.028) &0.47 (0.035) &1.08 (0.049) \\
		PMF &0.12 (0.097) &0.56 (0.065) &0.90 (0.039) &1.36 (0.067) &0.02 (0.015) &0.04 (0.023) &0.17 (0.034) &0.47 (0.059) \\
		LMF &0.36 (0.168) &0.56 (0.282) &0.95 (0.237) &1.29 (0.154) &0.02 (0.011) &0.10 (0.044) &0.17 (0.062) &0.50 (0.058) \\ \hline
		EMF &1.43 (0.292) &1.51 (0.343) &2.62 (0.383) &2.79 (0.243) &0.14 (0.032) &0.16 (0.036) &0.34 (0.067) &0.69 (0.080) \\
		SMF &\textbf{4.52 (1.029)} &\textbf{5.56 (0.394)} &\textbf{5.67 (0.433)} &\textbf{5.74 (0.159)} &\textbf{0.29 (0.064)} &\textbf{0.49 (0.038)} &\textbf{0.74 (0.062)} &\textbf{1.51 (0.049)} \\ \hline
	\end{tabular}
	\caption{Precision and Recall (mean and standard error) on the OTT dataset for $K = 20$.} 
	\label{tab:ott_20_tab}
\end{table*}

\begin{table*}[t] 
	\centering
	\scriptsize
	\begin{tabular}{l||c|c|c|c||c|c|c|c}
		\hline
		\multirow{2}{*}{\textbf{Method}} &
		\multicolumn{4}{c||}{\textbf{Precision}} & \multicolumn{4}{c}{\textbf{Recall}}\\
		& \multicolumn{1}{c|}{@2 $\left[\times10^{-2}\right]$} & \multicolumn{1}{c|}{@3 $\left[\times10^{-2}\right]$}& \multicolumn{1}{c|}{@5 $\left[\times10^{-2}\right]$}& \multicolumn{1}{c||}{@10 $\left[\times10^{-2}\right]$}&
		\multicolumn{1}{c|}{@2 $\left[\times10^{-2}\right]$} & \multicolumn{1}{c|}{@3 $\left[\times10^{-2}\right]$}& \multicolumn{1}{c|}{@5 $\left[\times10^{-2}\right]$}& \multicolumn{1}{c}{@10 $\left[\times10^{-2}\right]$}\\
		\hline 
		MF &2.94 (0.000) &4.31 (0.640) &5.18 (0.293) &6.31 (0.115) &0.01 (0.001) &0.01 (0.002) &0.02 (0.002) &0.05 (0.002) \\
		NMF &0.00 (0.000) &0.00 (0.000) &0.08 (0.064) &0.20 (0.064) &0.00 (0.000) &0.00 (0.000) &0.00 (0.000) &0.00 (0.000) \\
		BMF &0.78 (0.424) &0.52 (0.282) &0.55 (0.169) &1.06 (0.192) &0.00 (0.001) &0.00 (0.001) &0.00 (0.001) &0.01 (0.003) \\
		PMF &1.57 (0.577) &1.96 (0.489) &2.98 (0.420) &4.31 (0.401) &0.02 (0.006) &0.02 (0.008) &0.04 (0.013) &0.09 (0.015) \\
		LMF &3.14 (0.160) &3.01 (0.107) &4.08 (0.231) &6.59 (0.508) &0.01 (0.003) &0.03 (0.002) &0.06 (0.013) &0.16 (0.032) \\ \hline
		EMF &3.14 (0.424) &5.10 (0.320) &5.88 (0.111) &6.63 (0.115) &0.01 (0.001) &0.01 (0.001) &0.02 (0.002) &0.05 (0.001) \\
		SMF &\textbf{7.06 (0.277)} &\textbf{7.58 (0.427)} &\textbf{8.78 (0.279)} &\textbf{9.37 (0.210)} &\textbf{0.04 (0.009)} &\textbf{0.05 (0.013)} &\textbf{0.10 (0.008)} &\textbf{0.16 (0.009)} \\ \hline
	\end{tabular}
	\caption{Precision and Recall (mean and standard error) on the CTR dataset for $K = 10$.} 
	\label{tab:ctr_10_tab}
\end{table*}
\begin{table*}[t] 
	\centering
	\scriptsize
	\begin{tabular}{l||c|c|c|c||c|c|c|c}
		\hline
		\multirow{2}{*}{\textbf{Method}} &
		\multicolumn{4}{c||}{\textbf{Precision}} & \multicolumn{4}{c}{\textbf{Recall}}\\
		& \multicolumn{1}{c|}{@2 $\left[\times10^{-2}\right]$} & \multicolumn{1}{c|}{@3 $\left[\times10^{-2}\right]$}& \multicolumn{1}{c|}{@5 $\left[\times10^{-2}\right]$}& \multicolumn{1}{c||}{@10 $\left[\times10^{-2}\right]$}&
		\multicolumn{1}{c|}{@2 $\left[\times10^{-2}\right]$} & \multicolumn{1}{c|}{@3 $\left[\times10^{-2}\right]$}& \multicolumn{1}{c|}{@5 $\left[\times10^{-2}\right]$}& \multicolumn{1}{c}{@10 $\left[\times10^{-2}\right]$}\\
		\hline 
		MF &2.94 (0.277) &3.79 (0.107) &5.02 (0.231) &6.35 (0.096) &0.00 (0.001) &0.01 (0.001) &0.02 (0.001) &0.05 (0.002) \\
		NMF &0.00 (0.000) &0.00 (0.000) &0.00 (0.000) &0.24 (0.055) &0.00 (0.000) &0.00 (0.000) &0.00 (0.000) &0.00 (0.001) \\
		BMF &1.18 (0.277) &1.96 (0.370) &2.12 (0.222) &2.47 (0.242) &0.00 (0.001) &0.00 (0.001) &0.01 (0.001) &0.02 (0.002) \\
		PMF &2.94 (0.555) &3.53 (0.185) &3.45 (0.128) &4.78 (0.547) &0.01 (0.001) &0.01 (0.001) &0.03 (0.007) &0.10 (0.025) \\
		LMF &3.53 (0.000) &3.66 (0.213) &4.55 (0.256) &7.18 (0.640) &0.02 (0.007) &0.04 (0.006) &0.06 (0.013) &0.15 (0.033) \\ \hline
		EMF &3.14 (0.424) &5.10 (0.320) &5.80 (0.064) &6.75 (0.032) &0.01 (0.001) &0.01 (0.001) &0.02 (0.001) &0.05 (0.003) \\
		SMF &\textbf{8.24 (0.000)} &\textbf{9.15 (0.282)} &\textbf{9.96 (0.279)} &\textbf{11.10 (0.378)} &\textbf{0.04 (0.003)} &\textbf{0.05 (0.007)} &\textbf{0.09 (0.011)} &\textbf{0.17 (0.024)} \\ \hline
		
	\end{tabular}
	\caption{Precision and Recall (mean and standard error) on the CTR dataset for $K = 15$.} 
	\label{tab:ctr_15_tab}
\end{table*}
\begin{table*}[t] 
	\centering
	\scriptsize
	\begin{tabular}{l||c|c|c|c||c|c|c|c}
		\hline
		\multirow{2}{*}{\textbf{Method}} &
		\multicolumn{4}{c||}{\textbf{Precision}} & \multicolumn{4}{c}{\textbf{Recall}}\\
		& \multicolumn{1}{c|}{@2 $\left[\times10^{-2}\right]$} & \multicolumn{1}{c|}{@3 $\left[\times10^{-2}\right]$}& \multicolumn{1}{c|}{@5 $\left[\times10^{-2}\right]$}& \multicolumn{1}{c||}{@10 $\left[\times10^{-2}\right]$}&
		\multicolumn{1}{c|}{@2 $\left[\times10^{-2}\right]$} & \multicolumn{1}{c|}{@3 $\left[\times10^{-2}\right]$}& \multicolumn{1}{c|}{@5 $\left[\times10^{-2}\right]$}& \multicolumn{1}{c}{@10 $\left[\times10^{-2}\right]$}\\
		\hline 
		MF &2.75 (0.160) &3.40 (0.213) &5.65 (0.192) &6.59 (0.055) &0.00 (0.001) &0.01 (0.001) &0.02 (0.002) &0.05 (0.003) \\
		NMF &0.00 (0.000) &0.00 (0.000) &0.00 (0.000) &0.20 (0.032) &0.00 (0.000) &0.00 (0.000) &0.00 (0.000) &0.00 (0.000) \\
		BMF &1.96 (0.320) &1.83 (0.213) &2.35 (0.293) &3.25 (0.390) &0.00 (0.001) &0.00 (0.000) &0.00 (0.001) &0.02 (0.001) \\
		PMF &2.35 (0.000) &3.40 (0.700) &4.31 (0.128) &4.71 (0.364) &0.02 (0.007) &0.03 (0.008) &0.04 (0.007) &0.07 (0.012) \\
		LMF &2.94 (0.000) &3.66 (0.213) &5.25 (0.279) &7.88 (0.454) &0.01 (0.006) &0.04 (0.006) &0.06 (0.012) &0.17 (0.032) \\ \hline
		EMF &3.14 (0.424) &5.10 (0.320) &5.80 (0.064) &6.63 (0.115) &0.01 (0.001) &0.01 (0.001) &0.02 (0.001) &0.05 (0.001) \\
		SMF &\textbf{9.41 (0.734)} &\textbf{9.67 (0.565)} &\textbf{10.20 (0.256)} &\textbf{10.16 (0.250)} &\textbf{0.06 (0.007)} &\textbf{0.07 (0.006)} &\textbf{0.12 (0.013)} &\textbf{0.19 (0.016)} \\ \hline
	\end{tabular}
	\caption{Precision and Recall (mean and standard error) on the CTR dataset for $K = 20$.} 
	\label{tab:ctr_20_tab}
\end{table*}

\clearpage
\bibliographystyle{named}
\bibliography{bib}

\begin{thebibliography}{}

\bibitem[\protect\citeauthoryear{Aharon \bgroup \em et al.\egroup
  }{2006}]{aharon2006k}
Michal Aharon, Michael Elad, Alfred Bruckstein, et~al.
\newblock K-svd: An algorithm for designing overcomplete dictionaries for
  sparse representation.
\newblock {\em IEEE Transactions on signal processing}, 54(11):4311, 2006.

\bibitem[\protect\citeauthoryear{Arias \bgroup \em et al.\egroup
  }{2018}]{arias2018human}
Michael Arias, Rodrigo Saavedra, Maira~R Marques, Jorge Munoz-Gama, and Marcos
  Sep{\'u}lveda.
\newblock Human resource allocation in business process management and process
  mining: A systematic mapping study.
\newblock {\em Management Decision}, 56(2):376--405, 2018.

\bibitem[\protect\citeauthoryear{Boyd and Vandenberghe}{2004}]{boyd2004convex}
Stephen Boyd and Lieven Vandenberghe.
\newblock {\em Convex optimization}.
\newblock Cambridge university press, 2004.

\bibitem[\protect\citeauthoryear{Cand{\`e}s and Plan}{2010}]{candes2010matrix}
Emmanuel~J Cand{\`e}s and Yaniv Plan.
\newblock Matrix completion with noise.
\newblock {\em Proceedings of the IEEE}, 98(6):925--936, 2010.

\bibitem[\protect\citeauthoryear{Cand{\`e}s and Recht}{2009}]{candes2009exact}
Emmanuel~J Cand{\`e}s and Benjamin Recht.
\newblock Exact matrix completion via convex optimization.
\newblock {\em Foundations of Computational mathematics}, 9(6):717, 2009.

\bibitem[\protect\citeauthoryear{Conforti \bgroup \em et al.\egroup
  }{2015}]{conforti2015recommendation}
Raffaele Conforti, Massimiliano de~Leoni, Marcello La~Rosa, Wil~MP van~der
  Aalst, and Arthur~HM ter Hofstede.
\newblock A recommendation system for predicting risks across multiple business
  process instances.
\newblock {\em Decision Support Systems}, 69:1--19, 2015.

\bibitem[\protect\citeauthoryear{Dumas \bgroup \em et al.\egroup
  }{2013}]{dumas2013fundamentals}
Marlon Dumas, Marcello La~Rosa, Jan Mendling, Hajo~A Reijers, et~al.
\newblock {\em Fundamentals of business process management}, volume~1.
\newblock Springer, 2013.

\bibitem[\protect\citeauthoryear{Fang \bgroup \em et al.\egroup
  }{2017}]{Fang:2017:IBM:3172077.3172117}
Huang Fang, Zhen Zhang, Yiqun Shao, and Cho-Jui Hsieh.
\newblock Improved bounded matrix completion for large-scale recommender
  systems.
\newblock In {\em Proceedings of the 26th International Joint Conference on
  Artificial Intelligence}, IJCAI'17, pages 1654--1660. AAAI Press, 2017.

\bibitem[\protect\citeauthoryear{Fu \bgroup \em et al.\egroup
  }{2018}]{fu2018identifiability}
Xiao Fu, Kejun Huang, and Nicholas~D Sidiropoulos.
\newblock On identifiability of nonnegative matrix factorization.
\newblock {\em IEEE Signal Processing Letters}, 25(3):328--332, 2018.

\bibitem[\protect\citeauthoryear{Funk}{2011}]{funk2011netflix}
Simon Funk.
\newblock Netflix update: Try this at home, 2006, 2011.

\bibitem[\protect\citeauthoryear{Hoeffding}{1963}]{hoeffding1963probability}
Wassily Hoeffding.
\newblock Probability inequalities for sums of bounded random variables.
\newblock {\em Journal of the American statistical association},
  58(301):13--30, 1963.

\bibitem[\protect\citeauthoryear{Hoyer}{2004}]{hoyer2004non}
Patrik~O Hoyer.
\newblock Non-negative matrix factorization with sparseness constraints.
\newblock {\em Journal of machine learning research}, 5(Nov):1457--1469, 2004.

\bibitem[\protect\citeauthoryear{Hu \bgroup \em et al.\egroup
  }{2008}]{hu2008collaborative}
Yifan Hu, Yehuda Koren, and Chris Volinsky.
\newblock Collaborative filtering for implicit feedback datasets.
\newblock In {\em Data Mining, 2008. ICDM'08. Eighth IEEE International
  Conference on}, pages 263--272. Ieee, 2008.

\bibitem[\protect\citeauthoryear{Huang \bgroup \em et al.\egroup
  }{2012}]{huang2012task}
Zhengxing Huang, Xudong Lu, and Huilong Duan.
\newblock A task operation model for resource allocation optimization in
  business process management.
\newblock {\em IEEE Transactions on Systems, man, and cybernetics-part a:
  systems and humans}, 42(5):1256--1270, 2012.

\bibitem[\protect\citeauthoryear{Jain \bgroup \em et al.\egroup
  }{2013}]{jain2013low}
Prateek Jain, Praneeth Netrapalli, and Sujay Sanghavi.
\newblock Low-rank matrix completion using alternating minimization.
\newblock In {\em Proceedings of the forty-fifth annual ACM symposium on Theory
  of computing}, pages 665--674. ACM, 2013.

\bibitem[\protect\citeauthoryear{Jawanpuria and
  Mishra}{2018}]{jawanpuria2018unified}
Pratik Jawanpuria and Bamdev Mishra.
\newblock A unified framework for structured low-rank matrix learning.
\newblock In {\em International Conference on Machine Learning}, pages
  2259--2268, 2018.

\bibitem[\protect\citeauthoryear{Jiang \bgroup \em et al.\egroup
  }{2018}]{jiang2018magnitude}
Shuai Jiang, Kan Li, Da~Xu, and Richard Yi.
\newblock Magnitude bounded matrix factorisation for recommender systems.
\newblock {\em arXiv preprint arXiv:1807.05515}, 2018.

\bibitem[\protect\citeauthoryear{Johnson}{2014}]{johnson2014logistic}
Christopher~C Johnson.
\newblock Logistic matrix factorization for implicit feedback data.
\newblock 2014.

\bibitem[\protect\citeauthoryear{Kannan \bgroup \em et al.\egroup
  }{2014}]{kannan2014bounded}
Ramakrishnan Kannan, Mariya Ishteva, and Haesun Park.
\newblock Bounded matrix factorization for recommender system.
\newblock {\em Knowledge and information systems}, 39(3):491--511, 2014.

\bibitem[\protect\citeauthoryear{Koren \bgroup \em et al.\egroup
  }{2009}]{koren2009matrix}
Yehuda Koren, Robert Bell, and Chris Volinsky.
\newblock Matrix factorization techniques for recommender systems.
\newblock {\em Computer}, 42(8):30--37, 2009.

\bibitem[\protect\citeauthoryear{Lin \bgroup \em et al.\egroup
  }{2015}]{lin2015identifiability}
Chia-Hsiang Lin, Wing-Kin Ma, Wei-Chiang Li, Chong-Yung Chi, and ArulMurugan
  Ambikapathi.
\newblock Identifiability of the simplex volume minimization criterion for
  blind hyperspectral unmixing: The no-pure-pixel case.
\newblock {\em IEEE Transactions on Geoscience and Remote Sensing},
  53(10):5530--5546, 2015.

\bibitem[\protect\citeauthoryear{Mnih and
  Salakhutdinov}{2008}]{mnih2008probabilistic}
Andriy Mnih and Ruslan~R Salakhutdinov.
\newblock Probabilistic matrix factorization.
\newblock In {\em Advances in neural information processing systems}, pages
  1257--1264, 2008.

\bibitem[\protect\citeauthoryear{Salakhutdinov and
  Mnih}{2008}]{Salakhutdinov:2008:BPM:1390156.1390267}
Ruslan Salakhutdinov and Andriy Mnih.
\newblock Bayesian probabilistic matrix factorization using markov chain monte
  carlo.
\newblock In {\em Proceedings of the 25th International Conference on Machine
  Learning}, ICML '08, pages 880--887, New York, NY, USA, 2008. ACM.

\bibitem[\protect\citeauthoryear{Soni \bgroup \em et al.\egroup
  }{2016}]{soni2016noisy}
Akshay Soni, Swayambhoo Jain, Jarvis Haupt, and Stefano Gonella.
\newblock Noisy matrix completion under sparse factor models.
\newblock {\em IEEE Transactions on Information Theory}, 62(6):3636--3661,
  2016.

\bibitem[\protect\citeauthoryear{Van Der~Aalst}{2011}]{van2011process}
Wil Van Der~Aalst.
\newblock {\em Process mining: discovery, conformance and enhancement of
  business processes}, volume~2.
\newblock Springer, 2011.

\bibitem[\protect\citeauthoryear{Wang and
  Carreira-Perpin{\'a}n}{2013}]{wang2013projection}
Weiran Wang and Miguel~A Carreira-Perpin{\'a}n.
\newblock Projection onto the probability simplex: An efficient algorithm with
  a simple proof, and an application.
\newblock {\em arXiv preprint arXiv:1309.1541}, 2013.

\end{thebibliography}

\end{document}